\documentclass{article}

\usepackage{arxiv}
\usepackage[utf8]{inputenc}
\usepackage[T1]{fontenc}    
\usepackage{hyperref}       
\hypersetup{
    colorlinks=false,
    pdfborder={0 0 0}, 
}
\usepackage{amsmath}
\usepackage{url}            
\usepackage{booktabs}       
\usepackage{amsfonts}       
\usepackage{nicefrac}       
\usepackage{microtype}      
\usepackage{lipsum}		
\usepackage{graphicx}
\usepackage{doi}
\usepackage[square, numbers]{natbib}

\makeatletter
\newenvironment{algorithm}{%
  \par\list{}{\leftmargin\z@ \labelwidth\z@ \labelsep\z@
              \itemsep6pt \topsep12pt plus2pt
              }%
}{\endlist}

\makeatother

\title{Computational Social Linguistics for Telugu Cultural Preservation: 
Novel Algorithms for Chandassu Metrical Pattern Recognition}

\makeatletter
\renewcommand{\@title}{\text{Computational} Social Linguistics for Telugu Cultural Preservation: 
Novel Algorithms for Chandassu Metrical Pattern Recognition}
\makeatother

\author{ \href{https://orcid.org/0009-0003-3233-9447}{\includegraphics[scale=0.06]{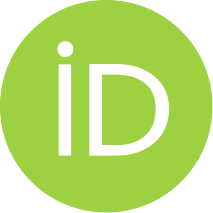}\hspace{1mm}Boddu Sri Pavan} \\
Independent Researcher \\
        \texttt{boddusripavan111@gmail.com} \\
    \And
    \href{https://orcid.org/0000-0001-7489-1174}{\includegraphics[scale=0.06]{orcid.pdf}\hspace{1mm}Chandrasheker Thummanagoti} \\
     Department of Computer Science and Engineering\\
    RGUKT, Nuzvid \\
	\texttt{chandra.indra@rguktn.ac.in} \\
	\And
	\href{https://orcid.org/0009-0000-3793-7743}{\includegraphics[scale=0.06]{orcid.pdf}\hspace{1mm}Boddu Swathi Sree} \\
    Department of Computer Science and Engineering\\
    RGUKT, Nuzvid \\
	\texttt{n220962@rguktn.ac.in} \\
}

\date{}

\renewcommand{\headeright}{}
\renewcommand{\undertitle}{A Preprint}
\renewcommand{\shorttitle}{Computational Social Linguistics for Telugu Cultural Preservation: Novel Algorithms for Chandassu Metrical Pattern Recognition}

\hypersetup{
pdftitle={Computational Social Linguistics for Telugu Cultural Preservation: Novel Algorithms for Chandassu Metrical Pattern Recognition},
pdfsubject={cs.CL},
pdfauthor={Boddu Sri Pavan, Boddu Swathi Sree},
pdfkeywords={Computational Social Linguistics, Natural Language Processing, Computational Linguistics, Prosodic Pattern Recognition, AksharamTokenizer, Chandassu Score }
}

\begin{document}

\maketitle

\begin{abstract}

\textit{This research presents a computational social science approach to preserving Telugu Chandassu, the metrical poetry tradition representing centuries of collective cultural intelligence. We develop the first comprehensive digital framework for analyzing Telugu prosodic patterns, bridging traditional community knowledge with modern computational methods. Our social computing approach involves collaborative dataset creation of 4,651 annotated padyams, validated linguistic patterns, and culturally informed algorithmic design. The framework includes AksharamTokenizer for prosody-aware tokenization, LaghuvuGuruvu Generator for classifying light and heavy syllables, and PadyaBhedam Checker for automated pattern recognition. Our algorithm achieves 91.73\% accuracy on the proposed Chandassu Score, with evaluation metrics reflecting traditional literary standards. This work demonstrates how computational social science can preserve endangered cultural knowledge systems while enabling new forms of collective intelligence around literary heritage. The methodology offers insights for community-centered approaches to cultural preservation, supporting broader initiatives in digital humanities and socially-aware computing systems.}

\end{abstract}

\keywords{Computational Social Linguistics \and Natural Language Processing \and Computational Linguistics \and Prosodic Pattern Recognition \and AksharamTokenizer \and Chandassu Score}

\section{Introduction}

\setlength{\parindent}{0pt}Telugu, an Indian language spoken by over 75 million people \cite{telugu_population}, possesses a rich tradition of metrical poetry known as chandassu. Unlike free verse, chandassu poetry adheres to strict prosodic patterns that govern syllabic weight, rhythm, and structural organization. The computational analysis of such metrical systems presents unique challenges due to the complex interplay between phonological properties, orthographic representation, and traditional prosodic rules. The fundamental units of Telugu prosody are laghuvu (light syllables) and guruvu (heavy syllables), traditionally represented by the symbols "|" and "U" respectively. Every aksharam (the smallest human-perceivable unit in the Telugu script) is classified as either laghuvu or guruvu based on its phonological duration and adherence to classical Telugu prosodic rules \cite{chando_vyakaranamu, pedda_bala_siksha}. These binary classifications form the foundation for higher-level metrical structures.

\setlength{\parindent}{15pt} Ganams represent sequential patterns of laghuvu-guruvu combinations that serve as the building blocks of metrical verses. Based on the number of aksharam tokens contained in each ganam, ganams are categorized into four groups, as shown in Table~\ref{tab:cat_ganam_at}. This taxonomy encompasses the complete spectrum from minimal single-aksharam ganams (LA: |, GA: U) to complex four-aksharam patterns (NALA: ||||, NAGA: |||U, SALA: ||U|).

\begin{table}[b!]
\centering
\begin{tabular}{lll}
\toprule
Aksharam Token Count& Ganam Name& LaghuvuGuruvu Pattern\\
\midrule
1& LA& | \\
1& GA& U \\
2& LAA& || \\
2& VA& |U \\
2& HA& U| \\
2& GAA& UU \\
3& NA& ||| \\
3& SA& ||U \\
3& JA& |U| \\
3& YA& |UU \\
3& BHA& U|| \\
3& RA& U|U \\
3& THA& UU| \\
3& MA& UUU \\
4& NALA& |||| \\
4& NAGA& |||U \\
4& SALA& ||U| \\
\bottomrule
\end{tabular}
\vspace{0.6em}
\caption{Ganam classification by aksharam token count and laghuvu-guruvu pattern}
\label{tab:cat_ganam_at}
\end{table}

Telugu prosody further categorizes ganams into two functional classes: \textit{Indra Ganams} (BHA, RA, THA, NALA, NAGA, SALA) and \textit{Surya Ganams} (HA, NA). This classification reflects traditional pedagogical approaches and influences the permissible substitutions within metrical patterns.

Telugu metrical poetry encompasses three primary structural classes, each containing multiple varieties (types) with distinct constraints, as described in Table~\ref{tab:chandassu_dataset}. These constraints govern various aspects including the number of paadams (lines), aksharam count per each paadam (number of aksharam tokens per each line), ganam sequences, yati (caesura), defined as the first aksharam token in each paadam, prasa (alliteration pattern), defined as the second aksharam token in each paadam, and prasa yati, a typical variant of yati. The interplay of these elements creates the complex metrical frameworks that define different poetic forms. Collections of such verses, particularly satakams (anthologies typically containing 100 or more padyams), represent significant literary achievements in Telugu tradition.

\begin{table}[b!]
\centering
\begin{tabular}{ll}
\toprule
Prosodic Class& Padyam Types\\
\midrule
Vruttamu& Vutpalamaala, Champakamaala, Saardulamu, Mattebhamu\\
Jaathi& Kandamu \\
Vupajaathi& Teytageethi, Aataveladi, Seesamu\\
\bottomrule
\vspace{0.6em}
\end{tabular}
\caption{Telugu prosodic classes and types in Chandassu dataset}
\label{tab:chandassu_dataset}
\end{table}

Despite the cultural and linguistic importance of chandassu, the comprehensive computational analysis of Telugu metrical poetry remains largely unexplored. Traditional manual analysis is time-intensive and requires extensive expertise in classical prosody. The lack of computational tools has hindered both preservation efforts and systematic analysis of large poetry corpora. Furthermore, the absence of standardized datasets and evaluation metrics has prevented the development of automated systems for metrical analysis and generation.

This work addresses these gaps by introducing the first comprehensive computational framework for Telugu chandassu analysis. Our contributions include: (1) a curated dataset of 4,651 padyams with LaghuvuGuruvu annotations, (2) AksharamTokenizer for prosodic analysis, (3) an algorithmic approach for automated chandassu identification, and (4) Chandassu Score as a quantitative evaluation metric. By providing these resources, we aim to advance both Telugu computational linguistics and the broader field of computational prosody, while contributing to the preservation and study of this rich literary tradition.

\section{Related Work}
\label{sec:Related Works}

Research in computational prosody has evolved from rule-based approaches for classical languages to modern machine learning techniques for contemporary poetry analysis. This section reviews relevant work in Telugu chandassu analysis, broader computational prosody, and evaluation methodologies.

\subsection{Telugu Tokenization Approaches}
Existing Telugu tokenization research has predominantly focused on word-level and subword-level segmentation for general NLP applications. SentencePiece \cite{kudo-richardson-2018-sentencepiece} and IndicNLP Suite \cite{kakwani-etal-2020-indicnlpsuite} provide comprehensive tokenization frameworks for Telugu, employing BPE, WordPiece \cite{wu2016googlesneuralmachinetranslation}, and rule-based approaches to handle morphological complexity, script normalization, and vocabulary construction for machine translation and language modeling tasks. Recent multilingual tokenization studies \cite{muennighoff-etal-2023-crosslingual} have evaluated Telugu tokenizer performance using efficiency metrics like Normalized Sequence Length across different encoding strategies, while approaches like \cite{karthika2025multilingualtokenizationlensindian} address agglutinative properties and syntactic boundaries in Telugu text processing. However, these existing approaches operate at granularities incompatible with prosodic analysis requirements. Conventional tokenizers optimize for computational efficiency and morpheme-based segmentation to support downstream NLP tasks, but fail to preserve the phonological properties essential for metrical pattern recognition.

\subsection{Traditional Telugu Prosody}
Classical Telugu prosody is comprehensively documented in foundational texts such as Chhan’doo vyaakarand-amu \cite{chando_vyakaranamu} and Telugu Vari Sampoorna Pedda Bala Siksha \cite{pedda_bala_siksha}. These works describe the theoretical framework for chandassu analysis, defining the constraints governing different padyam types, including ganam sequences, yati patterns, and prasa relationships. While these texts provide essential domain knowledge, they rely on manual analysis and lack computational formalization.

\subsection{Computational Prosody}
Early computational approaches to Telugu chandassu emerged in the 2010s. Rama N et al. \cite{rama_n} developed algorithms for metrical classification across multiple languages including Sanskrit and English, incorporating euphonic conjunction rules for verse correction. However, their approach assumes structural markers (punctuation) to identify paadams, limiting applicability to traditional Telugu texts where such markers are absent.

Raghu Korrapati et al. \cite{raghu_korrapati} introduced a natural language application for chandassu determination and word prediction in Telugu poetry. They categorized word complexity into three levels (simple, medium, complex) but did not provide quantitative evaluation metrics or systematic assessment of chandassu correctness. More recently, I. Reddy Sekhar Reddy, and M.Humera Khanam \cite{reddy_sekhar_reddy} proposed a Chandassu Recognizer (C.R.) system, while Nelli Ramesh, and S Vijay Kumar \cite{nelli_ramesh} developed VerseMetrics, a Python-based Telugu chandas analyzer. Both works claim accuracy improvements but lack detailed dataset descriptions, standardized evaluation metrics, and reproducible experimental protocols.

\subsection{Computational Prosody in Other Languages}
The broader field of computational prosody has seen significant advances in recent years. Navarro-Colorado B \cite{navarro_colorado} demonstrated topic modeling approaches for Spanish poetry analysis, while Thomas Haider \cite{haider2021metricaltaggingwildbuilding} developed universal tools for metrical tagging across multiple languages. For Indic traditions, Chandojñānam \cite{chandojnanam} identifies Sanskrit meters from laghu-guru patterns and uses fuzzy sequence matching to correct erroneous corpora. It illustrates cross-language use on Marathi and Telugu input but is not built around padyam-level constraints or a Telugu annotated benchmark. For Arabic, A. M. Mutawa1, and Ayshah Alrumaih \cite{arabicpoetry} report 97.53\% (full-verse) and 95.23\% (half-verse) accuracy with a bidirectional  long short-term memory model (LSTM) across 16 classical meters, using a held-out 15\% test split. For Sanskrit poetry generation, Manoj Balaji Jagadeeshan et al. \cite{sanskritgpt} measure metrical validity as syntactic accuracy, 99.86\% with constrained decoding versus 31.24\% (1-shot) for GPT-4o. For Russian, Ilya Koziev \cite{russiandataset} releases the RIFMA dataset consisting 5,121 samples of Russian-language poems (stanzas and whole poems) with stress marks and a rhyme scheme information. For Galician, Pablo Ruiz Fabo et al. \cite{pablo} annotate 4,287 examples and evaluate by exact per-line accuracy, exceeding 90\% for fine-tuned models against 89.9\% for a symbolic system.

These works highlight the importance of standardized datasets and robust evaluation frameworks- elements currently missing in Telugu chandassu research.

\subsection{Research Gaps}
Our analysis reveals several critical gaps in existing Telugu chandassu research:

\begin{enumerate}
\item \textbf{Dataset Limitations:} No publicly available, annotated datasets for Telugu metrical poetry analysis.
\item \textbf{Prosodic Tokenization:} The absence of specialised tokenizers human-perceivable, aksharam-level segmentation that preserve phonological weight properties essential for metrical analysis.
\item \textbf{Algorithmic Rigor:} Limited technical detail and reproducibility in existing approaches.
\item \textbf{Evaluation Metrics:} Absence of standardized quantitative measures for assessing chandassu correctness, which encompasses constraints such as yati, prasa, and ganam.
\item \textbf{Systematic Evaluation:} Lack of comprehensive experimental validation against established benchmarks.
\end{enumerate}
These gaps motivate our work's focus on dataset creation, specialized tokenizer design, algorithmic development, and standardized evaluation methodologies for Telugu computational prosody.

\section{Data Collection}
\label{sec:DataCollection}

\subsection{Data Acquisition}
We constructed the Chandassu dataset through systematic collection and rigorous validation of Telugu metrical poetry from digital archives. Our primary data source was: \url{https://andhrabharati.com/}, a comprehensive repository of classical Telugu literature containing digitized satakams (poetry collections) from various historical periods and authors. The data acquisition process employed targeted web scraping techniques to extract padyams from individual satakam pages, preserving the original text formatting and metadata. We developed custom scraping scripts that respected the website's structure while ensuring complete extraction of poetic content, including categorical classifications where available.

\subsection{Quality Assurance and Validation}
We implemented a two-stage validation process to ensure dataset integrity and annotation accuracy. In the first iteration, we manually verified each scraped padyam against its original source to ensure textual fidelity and correct Unicode representation. The second iteration occurred post-algorithm implementation, where we validated the automatically generated LaghuvuGuruvu sequences at the aksharam token level, verifying each prosodic annotation against classical Telugu metrical rules \cite{chando_vyakaranamu}, \cite{pedda_bala_siksha} to ensure accurate laghuvu-guruvu classification and successful pattern generation across all 4,651 padyams. The annotation and validation were carried out by the authors, who are native Telugu speakers. The authors studied Telugu poetry and chandassu from Secondary School Education at Zilla Parishad High School, Vetapalem through their undergraduate studies at Rajiv Gandhi University of Knowledge Technologies, Nuzvid.

\subsection{Dataset Structure and Composition}

The final Chandassu dataset comprises 4,651 carefully curated Telugu padyams spanning three primary prosodic classes and eight distinct types, as detailed in Table~\ref{tab:chandassu_dataset}. Each entry in the dataset contains the following structured attributes:

\begin{enumerate}
\item \textit{type:} type of padyam
\item \textit{padyam:} padyam text
\item \textit{class:} class of padyam
\item \textit{satakam:} satakam that the padyam belongs to
\item \textit{lg:} LaghuvuGuruvu data
\end{enumerate}

\begin{figure}[t!]
\begin{center}
\includegraphics[width=0.9\textwidth]{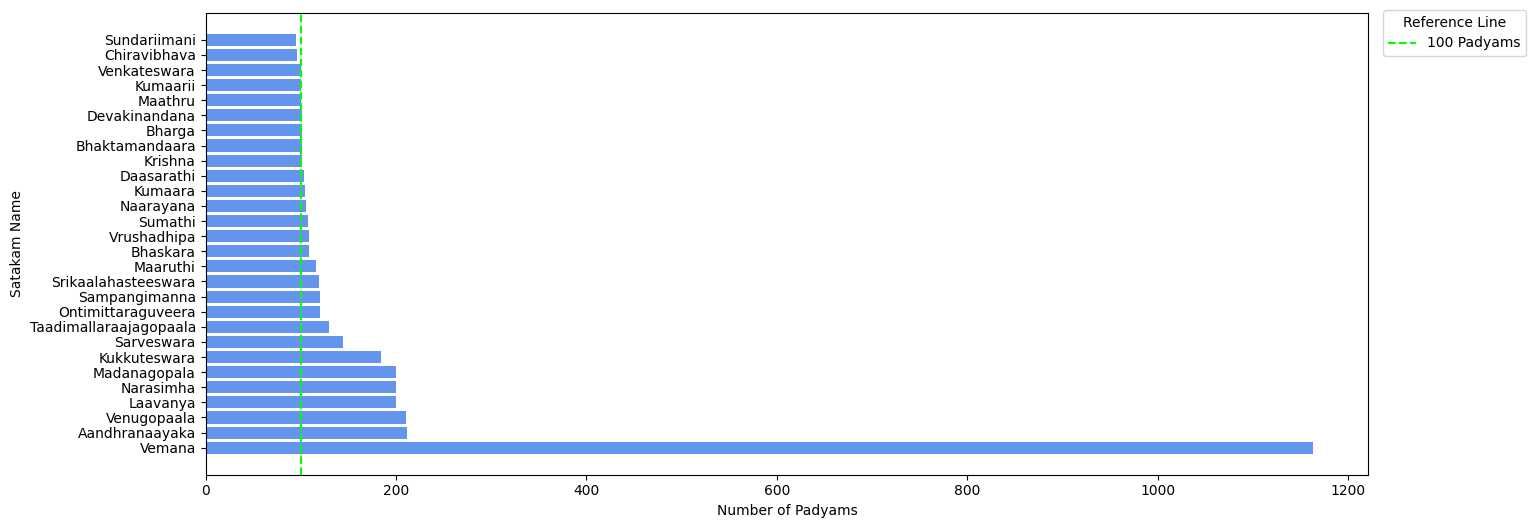}
\end{center}
\caption{Distribution of padyams across satakams in Chandassu dataset}
\label{fig:padyam_distribution_per_satakam}
\end{figure}

The distribution across padyam classes and types are presented in Table~\ref{tab:chandassu_dataset_class}, and Table~\ref{tab:chandassu_dataset_type}, respectively. These distributions reflect both the historical prevalence of different metrical forms and the availability of digitized texts. Aataveladi constitutes the largest category with 995 padyams, while Saardulamu contains the fewest with 290 padyams. Figure~\ref{fig:padyam_distribution_per_satakam} illustrates the distribution of padyams across individual satakams, revealing the diversity of sources contributing to our dataset and ensuring broad representation of Telugu metrical traditions.

\begin{table}[b!]
\centering
\begin{tabular}{ll}
\toprule
Prosodic Class& Padyam Count\\
\midrule
Vruttamu& 1625\\
Jaathi& 683\\
Vupajaathi& 2343\\
\bottomrule
\end{tabular}
\vspace{0.6em}
\caption{Distribution of padyams by prosodic class in Chandassu dataset}
\label{tab:chandassu_dataset_class}
\end{table}

\begin{table}[b!]
\centering
\begin{tabular}{ll}
\toprule
Padyam Type& Padyam Count\\
\midrule
Vutpalamaala& 329\\
Champakamaala& 389\\
Saardulamu& 290\\
Mattebhamu& 617\\
Kandamu& 683\\
Teytageethi& 676\\
Aataveladi& 995\\
Seesamu& 672\\
\bottomrule
\end{tabular}
\vspace{0.6em}
\caption{Distribution of padyams by type in Chandassu dataset}
\label{tab:chandassu_dataset_type}
\end{table}

Following algorithm development and evaluation (detailed in Section ~\ref{sec:methodology} and ~\ref{sec:evaluation_metrics}), we augmented the dataset with computed evaluation scores that quantify various aspects of metrical correctness. These augmented annotations transform the Chandassu dataset from a simple text corpus into a comprehensive evaluation benchmark for computational prosody research, enabling systematic assessment of algorithmic performance across multiple dimensions of metrical analysis.

To promote reproducibility and facilitate future research in Telugu computational poetry, we make the complete Chandassu dataset, and algorithm code  available at: \url{https://github.com/BodduSriPavan-111/chandassu}, under the MIT licence. This repository includes detailed documentation of collection procedures, validation protocols with reproducible Python code, ensuring transparent and reproducible research practices in this emerging field.

\section{Proposed Methodology}
\label{sec:methodology}

\begin{figure}[t!]
\begin{center}
\includegraphics[width=0.95\textwidth]{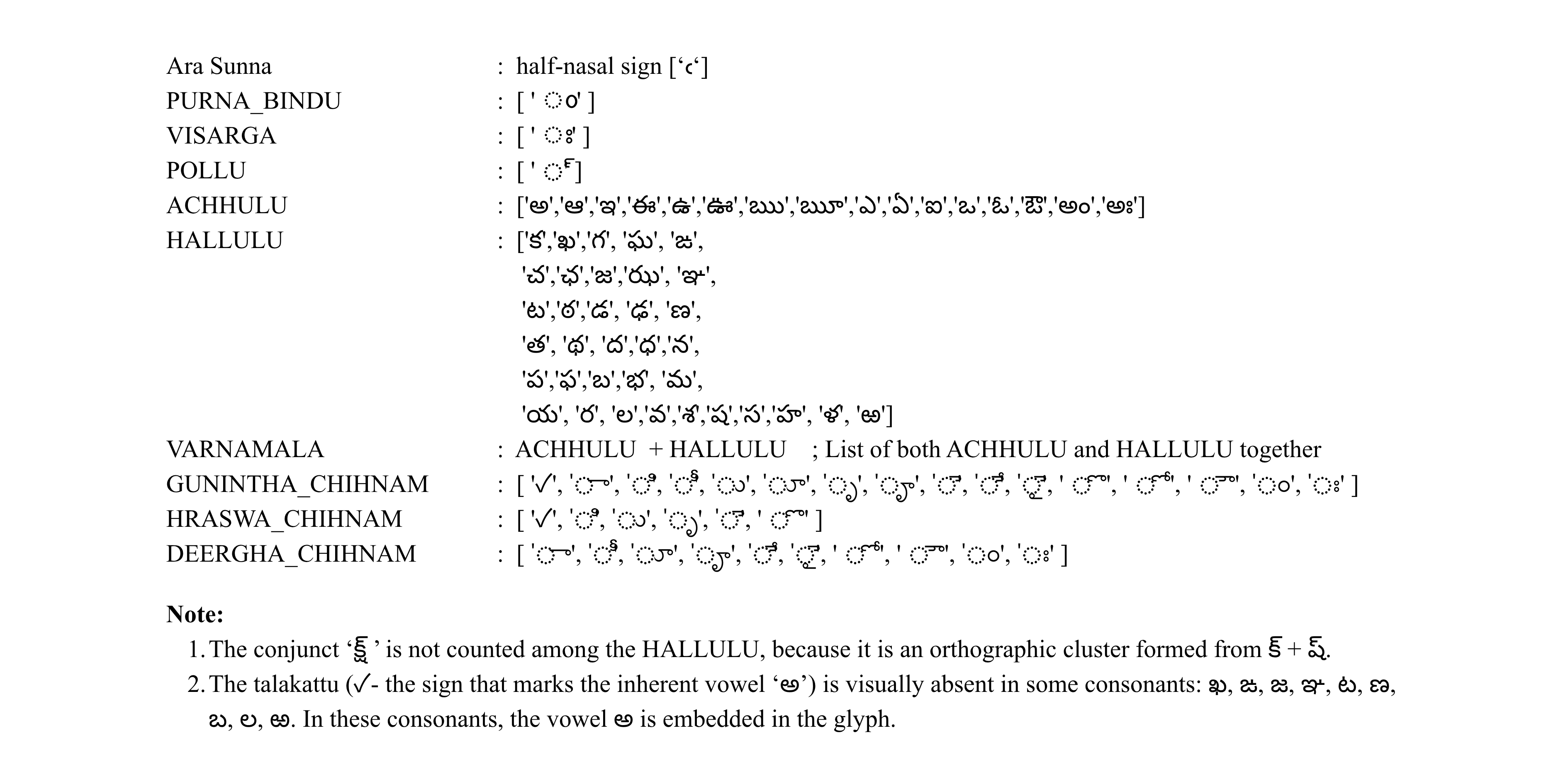}
\end{center}
\caption{Telugu character set used by the proposed framework}
\label{fig:character_set}
\end{figure}

\begin{figure}[t!]
\begin{center}
\includegraphics[width=0.85\textwidth]{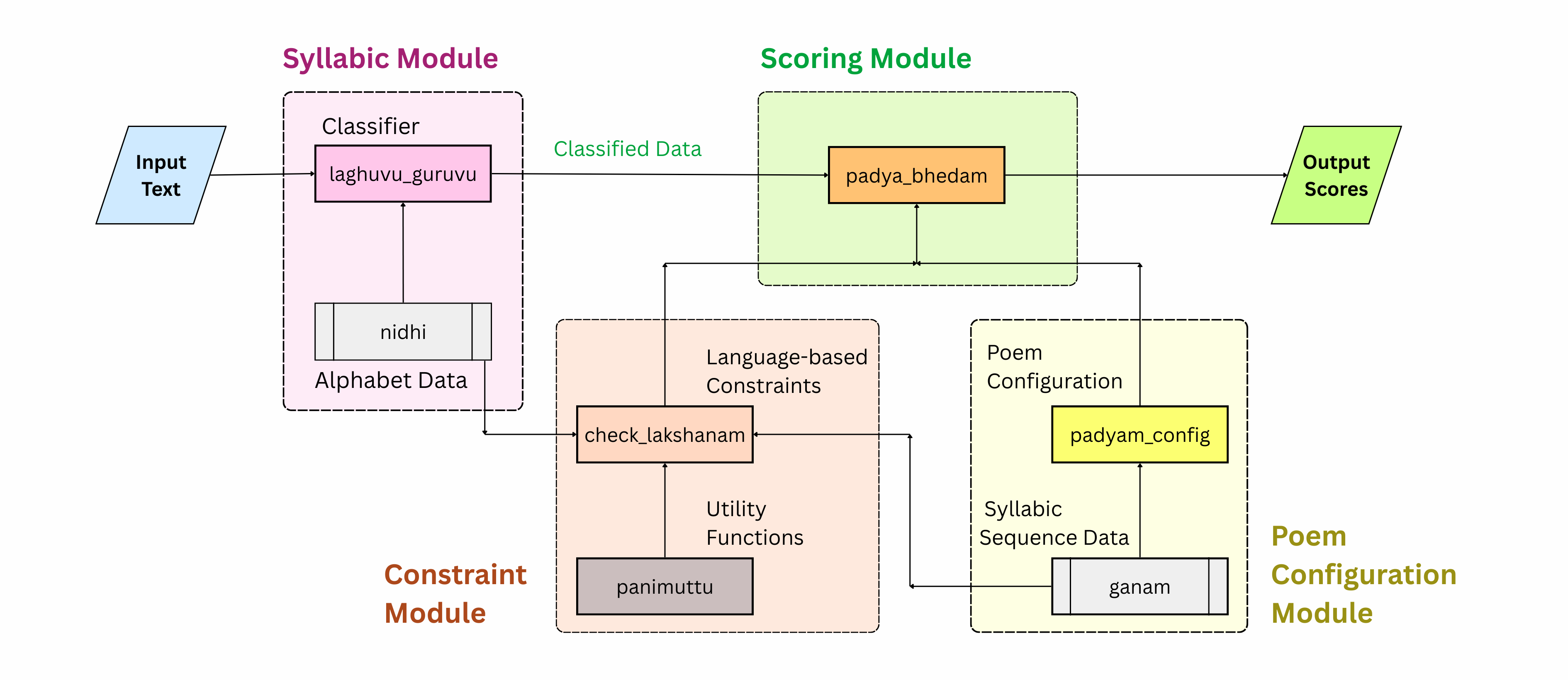}
\end{center}
\caption{Proposed computational framework architecture for Telugu metrical poetry analysis}
\label{fig:chandassu_architecture}
\end{figure}

Our computational framework for Telugu chandassu analysis employs a modular architecture that combines traditional prosodic knowledge with algorithmic processing. The system processes input padyams step-by-step: beginning with aksharam tokenization to identify human-perceivable character units, followed by prosodic pattern generation to determine laghuvu-guruvu sequences, and concluding with comprehensive metrical validation against predefined chandassu constraints. Figure~\ref{fig:character_set} contains values of all constants mentioned in the pseudocode. Figure~\ref{fig:chandassu_architecture} illustrates the complete pipeline architecture, comprising seven interconnected components that collectively enable automated chandassu verification and evaluation.

\begin{enumerate}
\item \textbf{nidhi}: Telugu character classification system containing the complete varnamala (alphabet) with phonological and orthographic categorizations essential for prosodic analysis.
\item \textbf{panimuttu}: Utility module providing specialized functions for vowel-consonant operations.
\item \textbf{laghuvu\_guruvu}: Prosodic analysis engine integrating the AksharamTokenizer and LaghuvuGuruvu generator for syllabic weight determination.
\item \textbf{ganam}: Ganam pattern repository containing mappings between LaghuvuGuruvu sequences and traditional metrical units.
\item \textbf{padyam\_config}: Configuration storing predefined constraints for each padyam type, including syllabic requirements, structural patterns, and prosodic rules.
\item \textbf{check\_lakshanam}: Constraint validation module for verifying adherence to traditional prosodic requirements such as yati (caesura) and prasa (alliteration).
\item \textbf{padya\_bhedam}: Integration module that orchestrates the evaluation process and generates comprehensive metrical assessments.
\end{enumerate}

\subsection{Aksharam Tokenizer}
The AksharamTokenizer addresses the fundamental challenge of identifying perceptually meaningful units in Telugu text for prosodic analysis. Unlike conventional tokenization approaches that operate at word or morpheme levels, our tokenizer extracts Aksharam Tokens- the smallest perceptually meaningful units used for metrical analysis in Telugu poetry. The tokenization process handles complex orthographic phenomena, including conjunct consonants (dwitwaaksharamulu, samyuktaaksharamulu), vowel diacritics, and grapheme clusters that are characteristic of Telugu script. Algorithm~\ref{algo:aksharamtokenizer} processes input text character by character, applying rule-based logic to group Unicode codepoints into coherent aksharam units while preserving their prosodic integrity. Word boundaries are identified through whitespace detection, enabling proper segmentation of paadams (poetic lines) essential for subsequent metrical analysis. 

We consider the character set shown in Figure \ref{fig:character_set} from \cite{pedda_bala_siksha}, \cite{chando_vyakaranamu}. Ara Sunna is a half-nasal sign found in archaic Telugu orthography. It does not influence the syllabic weight of the Aksharam Token \cite{pedda_bala_siksha}, \cite{chando_vyakaranamu}. Therefore, Step 7 of Algorithm \ref{algo:aksharamtokenizer} removes it before tokenization proceeds.

\begin{algorithm}
\item[Algorithm 4.1] AksharamTokenizer
\item[Input] INPUT\_TEXT $\in$ String - Telugu padyam text in Unicode format
\item[Output] List of prosodically segmented aksharam tokens
\item[Step 1] temp\_l= re.findall(r"\textbackslash X", INPUT\_TEXT)
\item[Step 2] l= Remove special characters, alphabets, and numbers from temp\_l
\item[Step 3] index= 0
\item[Step 4] text= []
\item[Step 5] temp= ""
\item[Step 6] \textbf{for} index from 0 to len(l) \textbf{do}
\item[Step 7] \quad l[index]= Remove Ara Sunna from l[index]
\item[Step 8] \quad \textbf{if} l[index].isspace \textbf{then}
\item[Step 9] \quad \quad \textbf{pass}
\item[Step 10] \quad \textbf{else if} l[index].endswith(POLLU) and temp == "" and index<(len(l)-1) and not l[index+1].isspace \textbf{then}
\item[Step 11] \quad \quad temp+= l[index]   
\item[Step 12] \quad \textbf{else if} l[index].endswith(POLLU) and temp != "" and not l[index+1].isspace \textbf{then}
\item[Step 13] \quad \quad temp+= l[index]
\item[Step 14] \quad \textbf{else if} l[index].endswith(POLLU) and (index+1 == len(l) or l[index+1].isspace ) \textbf{then}
\item[Step 15] \quad \quad text[-1]+= l[index] 
\item[Step 16] \quad \textbf{else if} (not l[index].endswith(POLLU)) and temp != "" \textbf{then}
\item[Step 17] \quad \quad text.append(temp+l[index])
\item[Step 18] \quad \quad temp= ""
\item[Step 19] \quad \textbf{else if} (not l[index].endswith(POLLU)) and temp == "" \textbf{then}
\item[Step 20] \quad \quad text.append(l[index])
\item[Step 21] \quad \quad temp= ""
\item[Step 22] \quad \textbf{end if}
\item[Step 23] \textbf{end for}
\item[Step 24] \textbf{return} text
\label{algo:aksharamtokenizer}
\end{algorithm}

\subsection{LaghuvuGuruvu Generator}

The LaghuvuGuruvu Generator, described in Algorithm~\ref{algo:laghuvuguruvu_generator}, transforms the sequence of aksharam tokens into prosodic representations by classifying each unit as either Laghuvu (light: |) or Guruvu (heavy: U) based on traditional Telugu prosodic rules \cite{chando_vyakaranamu,pedda_bala_siksha}. This binary classification forms the foundation for all subsequent metrical analysis and pattern matching.
The generation process incorporates phonological duration rules, orthographic conventions, and contextual factors that influence syllabic weight determination. The algorithm considers vowel length, consonant clusters, and conjunct formations that affect prosodic realization. Due to ambiguity in classical prosodic treatment of conjunct consonants containing the subscript "Ra" (Ra-vattu), we adopt a systematic approach of evaluating such aksharams without the subscript component to ensure consistent prosodic weight determination. In Algorithm \ref{algo:laghuvuguruvu_generator}, 'ACHHULU'  denotes the vowels and 'HALLULU' (Step 15) the consonants, and 'VARNAMALA' (Step 24) is the complete set formed by both, as shown in Figure \ref{fig:character_set}.

\begin{algorithm}
\item[Algorithm 4.2] LaghuvuGuruvuGenerator
\item[Input] INPUT\_TEXT $\in$ String - Telugu padyam text in Unicode format
\item[Output] List of aksharam tokens with corresponding laghuvu-guruvu annotations

\item[Step 1] l= AksharamTokenizer(INPUT\_TEXT)
\item[Step 2] marking= []
\item[Step 3] \textbf{for} index from 0 to len(l) \textbf{do}
\item[Step 4] \quad \textbf{if} index < (len(l)-1) \textbf{do}
\item[Step 5] \quad \quad x= re.findall(r"\textbackslash X", l[index+1])
\item[Step 6] \quad \quad \textbf{if} x[-1].endswith("POLLU") \textbf{then}
\item[Step 7] \quad \quad \quad x= "".join(x[:-1]) 
\item[Step 8] \quad \quad \textbf{else}
\item[Step 9] \quad \quad \quad x= "".join(x)
\item[Step 10] \quad \quad \textbf{end if}
\item[Step 11] \quad \quad d= frequency\_counter(x)
\item[Step 12] \quad \quad del d["Ra"]
\item[Step 13] \quad \quad temp\_count= 0
\item[Step 14] \quad \quad \textbf{for} i in d \textbf{do}
\item[Step 15] \quad \quad \quad \textbf{if} i in HALLULU \textbf{then}
\item[Step 16] \quad \quad \quad \quad temp\_count+= d[i]
\item[Step 17] \quad \quad \quad \textbf{end if}
\item[Step 18] \quad \quad \textbf{end for}
\item[Step 19] \quad \quad \textbf{if} "Ra" in l[index+1] and ((temp\_count==0) or (temp\_count == 1  and (not l[index+1].startswith("Ra")))) \textbf{then}
\item[Step 20] \quad \quad \quad marking.append(lg\_map[l[index][-1]])
\item[Step 21] \quad \quad \textbf{else}
\item[Step 22] \quad \quad \quad count= 0
\item[Step 23] \quad \quad \quad \textbf{for} j in list(l[index+1]) \textbf{do}
\item[Step 24] \quad \quad \quad \quad \textbf{if} j in VARNAMALA \textbf{then}
\item[Step 25] \quad \quad \quad \quad \quad count+= 1
\item[Step 26] \quad \quad \quad \quad \textbf{end if}
\item[Step 27] \quad \quad \quad \textbf{end for}
\item[Step 28] \quad \quad \quad \textbf{if} count > 1 and not l[index+1].endswith(POLLU) \textbf{then}
\item[Step 29] \quad \quad \quad \quad marking.append("U")
\item[Step 30] \quad \quad \quad \textbf{else if} count > 1 and l[index+1].endswith(POLLU) and re.findall(r"\textbackslash X", l[index+1])[0].endswith(POLLU) \textbf{then}
\item[Step 31] \quad \quad \quad \quad marking.append("U")
\item[Step 32] \quad \quad \quad \textbf{else if} count > 1 and l[index+1].endswith(POLLU) \textbf{then}
\item[Step 33] \quad \quad \quad \quad marking.append(lg\_map[l[index][-1]])
\item[Step 34] \quad \quad \quad \textbf{else}
\item[Step 35] \quad \quad \quad \quad marking.append(lg\_map[l[index][-1]])
\item[Step 36] \quad \quad \quad \textbf{end if}
\item[Step 37] \quad \quad \textbf{end if}
\item[Step 38] \quad \textbf{else if} l[index][-1] in lg\_map \textbf{then}
\item[Step 39] \quad \quad marking.append(lg\_map[l[index][-1]])
\item[Step 40] \quad \textbf{end if}
\item[Step 41] \textbf{end for}
\item[Step 42] \textbf{return} list(zip(l, marking))
\label{algo:laghuvuguruvu_generator}
\end{algorithm}

\subsection{PadyaBhedam Checker}

The PadyaBhedam Checker, described in Algorithm~\ref{algo:padya_bhedam}, serves as the primary evaluation engine, assessing generated LaghuvuGuruvu patterns against type-specific constraints stored in the configuration. In Algorithm 4.5 (PrasaYatiCheck), 'GUNINTHA\_CHIHNAM' (Step 20) denotes the vowel diacritics that modify the pronunciation of a consonant (HALLULU) when attached to it. 'HRASWA\_CHIHNAM' (Step 31) is the set of diacritics that make the consonant short in pronunciation, whereas 'DEERGHA\_CHIHNAM' (Step 35) is the set that makes it long, where both 'HRASWA\_CHIHNAM' and 'DEERGHA\_CHIHNAM' are subsets of 'GUNINTHA\_CHIHNAM'.

This module implements a comprehensive validation framework that examines multiple dimensions of metrical correctness. The checker evaluates structural conformity, including paadam count, aksharam count per line, and overall syllabic requirements. It performs ganam sequence matching to verify adherence to prescribed metrical patterns and validates yati placement according to traditional caesura rules. Additionally, the module assesses prasa, and prasa yati patterns to ensure proper alliterative relationships within and across paadams. Each evaluation dimension generates individual scores that contribute to the overall Chandassu Score, described in Equation~\ref{eq:chandassu_score}, providing granular feedback on metrical performance.

\begin{algorithm}
\item[Algorithm 4.3] PadyaBhedam

\item[Step 1] lg\_data= LaghuvuGuruvuGenerator(INPUT\_TEXT)
\item[Step 2] config= CONFIGURATION\_OF\_PADYAM to evaluate
\item[Step 3] ganamulu= GANAM\_TO\_SYMBOL\_MAP
\item[Step 4] r\_ganamulu= SYMBOL\_TO\_GANAM\_MAP
\item[Step 5] padamwise\_ganam\_data= []
\item[Step 6] gana\_kramam\_score= 0
\item[Step 7] end= 0
\item[Step 8] paadam\_count= 0
\item[Step 9] \textbf{for} line from 0 to len(config["gana\_kramam"])  \textbf{do}
\item[Step 10] \quad ganam\_data= []
\item[Step 11] \quad \textbf{for} j from 0 to len(config["gana\_kramam"][line])  \textbf{do}
\item[Step 12] \quad \quad ganam\_match\_flag= False
\item[Step 13] \quad \quad \textbf{for} i in config["gana\_kramam"][line][j]  \textbf{do}
\item[Step 14] \quad \quad \quad ganam= tuple([k[1] for k in lg\_data[end: end+len(ganamulu[i])]])
\item[Step 15] \quad \quad \quad  \textbf{if} r\_ganamulu[ganam] equals to i  \textbf{then}
\item[Step 16] \quad \quad \quad \quad ganam\_data.append([lg\_data[end:end+len(ganamulu[i])],r\_ganamulu[ganam]])
\item[Step 17] \quad \quad \quad \quad gana\_kramam\_score+= 1
\item[Step 18] \quad \quad \quad \quad ganam\_match\_flag= True
\item[Step 19] \quad \quad \quad \quad  \textbf{break}
\item[Step 20] \quad \quad \quad \textbf{else if} Exception occurred  \textbf{then}
\item[Step 21] \quad \quad \quad \quad pass
\item[Step 22] \quad \quad \quad \textbf{end if}
\item[Step 23] \quad \quad \textbf{end for}
\item[Step 24] \quad \quad \textbf{if} ganam\_match\_flag == False  \textbf{then}
\item[Step 25] \quad \quad \quad ganam\_data.append([lg\_data[end: end+len(ganamulu[i])], "UnMatched"])
\item[Step 26] \quad \quad  \textbf{end if}
\item[Step 27] \quad \quad end+= len(ganamulu[i])
\item[Step 28] \quad  \textbf{end for}
\item[Step 29] \quad  \textbf{if} len(ganam\_data[0][0])>1  \textbf{then}
\item[Step 30] \quad \quad paadam\_count+= 1
\item[Step 31] \quad  \textbf{end if}
\item[Step 32] \quad padamwise\_ganam\_data.append(ganam\_data)
\item[Step 33] \quad  \textbf{if} end>=len(lg\_data)  \textbf{then}
\item[Step 34] \quad \quad  \textbf{break}
\item[Step 35] \textbf{end for}
\item[Step 36] prasa\_yati\_match= PrasaYatiCheck(padamwise\_ganam\_data, type, config, config["only\_generic\_yati"])
\item[Step 37] N\_PAADALU= config["n\_paadalu"]
\item[Step 38] total\_yati\_paadalu= len(config["yati\_paadalu"])
\item[Step 39] \textbf{if} type== "seesamu"  \textbf{then}
\item[Step 40] \quad paadam\_count= paadam\_count/2
\item[Step 41] \textbf{end if}
\item[Step 42] score= \{"n\_paadalu\_score":  paadam\_count/ N\_PAADALU, "gana\_kramam\_score": gana\_kramam\_score/ sum([len(i) for i in config["gana\_kramam"]]), "yati\_score": sum(prasa\_yati\_match)/ total\_yati\_paadalu\}
\item[Step 43] \textbf{if} n\_aksharalu $\in$ config  \textbf{then}
\item[Step 44] \quad aksharam\_count= 0
\item[Step 45] \quad \textbf{for} i in padamwise\_ganam\_data  \textbf{do}
\item[Step 46] \quad \quad \textbf{for} j in i \textbf{do}
\item[Step 47] \quad \quad \quad aksharam\_count+= len(j[0])
\item[Step 48] \quad \quad \textbf{end for}
\item[Step 49] \quad \textbf{end for}
\item[Step 50] \quad subtract\_factor= len(lg\_data) - aksharam\_count
\item[Step 51] \quad score["n\_aksharaalu\_score"]= (aksharam\_count-subtract\_factor)/ (N\_PAADALU*config["n\_aksharalu"])
\item[Step 52] \textbf{end if}
\item[Step 53] \textbf{if} prasa $\in$ config \textbf{then}
\item[Step 54] \quad index= 2    \# Second aksharam token
\item[Step 55] \quad frequency= \{\}
\item[Step 56] \quad \textbf{for} i in padamwise\_ganam\_data \textbf{do}
\item[Step 57] \quad \quad \textbf{if} no Exception \textbf{then}
\item[Step 58] \quad \quad \quad aksharam= remove\_gunintha\_chihnam(i[0][0][index-1][0])
\item[Step 59] \quad \quad \quad frequency[aksharam]= frequency.get(aksharam,0)+ 1
\item[Step 60] \quad \textbf{end for}
\item[Step 61] \quad score["prasa\_score"]= max(frequency.values())/ N\_PAADALU
\item[Step 62] \textbf{end if}
\item[Step 63] overall\_score= sum(score.values())/ len(score)
\item[Step 64] \textbf{return} \{"chandassu\_score": overall\_score, "micro\_score": score\}
\label{algo:padya_bhedam}
\end{algorithm}

\begin{algorithm}
\item[Algorithm 4.4] YatiCheck

\item[Step 1] \textbf{if} first\_letter == None and yati\_sthanam\_letter == None \textbf{then}
\item[Step 2] \quad first\_letter= paadam[0]
\item[Step 3] \quad yati\_sthanam\_letter= paadam[yati\_sthanam-1]
\item[Step 4] \textbf{end if}
\item[Step 5] first\_letter= first\_letter.replace([PURNA\_BINDU, VISARGA], "")
\item[Step 6] yati\_sthanam\_letter= yati\_sthanam\_letter.replace([PURNA\_BINDU, VISARGA], "")
\item[Step 7] chihnam\_a= extract\_gunintha\_chihnam(first\_letter)
\item[Step 8] chihnam\_b= extract\_gunintha\_chihnam(yati\_sthanam\_letter)
\item[Step 9] chihna\_yati= False  \# Yati checking for ACHHULU (Vowels)
\item[Step 10] \textbf{for} i in yati \textbf{do}    
\item[Step 11] \quad \textbf{if} chihnam\_a in i \textbf{then}
\item[Step 12] \quad \quad \textbf{if} chihnam\_b in i \textbf{then}
\item[Step 13] \quad \quad \quad chihna\_yati= True
\item[Step 14] \quad \quad \textbf{else}
\item[Step 15] \quad \quad \quad \textbf{return} False
\item[Step 16] \quad \quad \textbf{end if}
\item[Step 17] \quad \textbf{end if}
\item[Step 18] \textbf{end for}
\item[Step 19] akshara\_yati= False \# Yati check for NON-ACCHULU (Non-vowels)
\item[Step 20] \textbf{for} i in list(set(extract\_aksharam(first\_letter))) \textbf{do}
\item[Step 21] \quad \textbf{for} j in yati \textbf{do}
\item[Step 22] \quad \quad \textbf{if} i in j \textbf{then}
\item[Step 23] \quad \quad \quad \textbf{for} k in list(set(extract\_aksharam(yati\_sthanam\_letter))) \textbf{do}
\item[Step 24] \quad \quad \quad \quad \textbf{if} k in j \textbf{then}
\item[Step 25] \quad \quad \quad \quad \quad akshara\_yati= True
\item[Step 26] \quad \quad \quad \quad \quad \textbf{break}
\item[Step 27] \quad \quad \quad \textbf{end for}
\item[Step 28] \quad \quad \quad \textbf{if} akshara\_yati == True \textbf{then}
\item[Step 29] \quad \quad \quad \quad \textbf{break}
\item[Step 30] \quad \quad \quad \textbf{end if}
\item[Step 31] \quad \quad \textbf{end if}
\item[Step 32] \quad \textbf{end for}
\item[Step 33] \textbf{end for}
\item[Step 34] \textbf{if} not akshara\_yati \textbf{then}
\item[Step 35] \quad \textbf{return} False
\item[Step 36] \textbf{end if}
\item[Step 37] \textbf{if} chihna\_yati and akshara\_yati \textbf{then}
\item[Step 38] \quad \textbf{return} True
\item[Step 39] \textbf{else}
\item[Step 40] \quad \textbf{return} False
\item[Step 41] \textbf{end if}
\item[Utility functions] extract\_gunintha\_chihnam extracts vowel diacritics, and extract\_aksharam extracts the alphabet from the given aksharam token.
\label{algo:yati_check}
\end{algorithm}

\begin{algorithm}
\item[Algorithm 4.5] PrasaYatiCheck

\item[Step 1] \textbf{if} type == "kandamu" \textbf{then}
\item[Step 2] \quad padamwise\_ganam\_data= [padamwise\_ganam\_data[i-1] for i in config["yati\_paadalu"]] 
\item[Step 3] \textbf{end if}
\item[Step 4] yati\_match= []
\item[Step 5] \textbf{for} row in padamwise\_ganam\_data \textbf{do}
\item[Step 6] \quad \textbf{if} (len(row[0][0])>1) and (len(row[config["yati\_sthanam"][0]-1][0])>1) \textbf{then}
\item[Step 7] \quad \quad first\_letter= [a[0] for a in row[0][0]]
\item[Step 8] \quad \quad yati\_sthanam\_letter= row[config["yati\_sthanam"][0]-1][0][config["yati\_sthanam"][1]]
\item[Step 9] \quad \quad generic\_yati= YatiCheck(first\_letter[0], yati\_sthanam\_letter[0])
\item[Step 10] \quad \quad \textbf{if} only\_generic\_yati == True \textbf{then}
\item[Step 11] \quad \quad \quad yati\_match.append(generic\_yati)
\item[Step 12] \quad \quad \quad \textbf{continue}
\item[Step 13] \quad \quad \textbf{end if}
\item[Step 14] \quad \quad yati\_sthanam\_letter= [a[0] for a in row[config["yati\_sthanam"][0]-1][0]][:2]
\item[Step 15] \quad \quad prasa\_yati\_match= False
\item[Step 16] \quad \quad \textbf{if} generic\_yati == True \textbf{then}
\item[Step 17] \quad \quad \quad yati\_match.append(True)
\item[Step 18] \quad \quad \textbf{else}
\item[Step 19] \quad \quad \quad hraswa\_deergham\_flag\_1= " "
\item[Step 20] \quad \quad \quad \textbf{if} first\_letter[0][-1] in GUNINTHA\_CHIHNAM \textbf{then}
\item[Step 21] \quad \quad \quad \quad hraswa\_deergham\_flag\_1= first\_letter[0][-1]
\item[Step 22] \quad \quad \quad \textbf{end if}
\item[Step 23] \quad \quad \quad hraswa\_deergham\_flag\_2= ""
\item[Step 24] \quad \quad \quad \textbf{if} yati\_sthanam\_letter[0][-1] in GUNINTHA\_CHIHNAM \textbf{then}
\item[Step 25] \quad \quad \quad \quad hraswa\_deergham\_flag\_2= yati\_sthanam\_letter[0][-1]
\item[Step 26] \quad \quad \quad \textbf{else}
\item[Step 27] \quad \quad \quad \quad hraswa\_deergham\_flag\_2= " "
\item[Step 28] \quad \quad \quad \textbf{end if}
\item[Step 29] \quad \quad \quad l1= remove\_gunintha\_chihnam(first\_letter[1])
\item[Step 30] \quad \quad \quad l2= remove\_gunintha\_chihnam(yati\_sthanam\_letter[1])
\item[Step 31] \quad \quad \quad \textbf{if} (hraswa\_deergham\_flag\_1 in HRASWA\_CHIHNAM and hraswa\_deergham\_flag\_2 in HRASWA\_CHIHNAM) \textbf{then}
\item[Step 32] \quad \quad \quad \quad \textbf{if}  l1 == l2 \textbf{then}
\item[Step 33] \quad \quad \quad \quad \quad prasa\_yati\_match= True
\item[Step 34] \quad \quad \quad \quad \textbf{end if}
\item[Step 35] \quad \quad \quad \textbf{else if} (hraswa\_deergham\_flag\_1 in DEERGHA\_CHIHNAM and hraswa\_deergham\_flag\_2 in DEERGHA\_CHIHNAM ) \textbf{then}
\item[Step 36] \quad \quad \quad \quad \textbf{if} l1 == l2 \textbf{then}
\item[Step 37] \quad \quad \quad \quad \quad prasa\_yati\_match= True
\item[Step 38] \quad \quad \quad \quad \textbf{end if}
\item[Step 39] \quad \quad \quad \textbf{end if}
\item[Step 40] \quad \quad \quad \textbf{if} prasa\_yati\_match \textbf{then}
\item[Step 41] \quad \quad \quad \quad yati\_match.append(True)
\item[Step 42] \quad \quad \quad \textbf{end if}
\item[Step 43] \quad \quad \textbf{end if}
\item[Step 44] \quad \quad \textbf{if} generic\_yati == False \textbf{and} prasa\_yati\_match == False \textbf{then}
\item[Step 45] \quad \quad \quad yati\_match.append(False)
\item[Step 46] \quad \quad \textbf{end if}
\item[Step 47] \quad \textbf{else}
\item[Step 48] \quad \quad yati\_match.append(False)
\item[Step 49] \quad \textbf{end if}  
\item[Step 50] \textbf{end for}
\item[Step 51] \textbf{return} yati\_match
\item[Note] 'config' defines the pre-defined configuration for each padyam type, 'remove\_gunintha\_chihnam' is a utility function to remove vowel diacritics from the given aksharam token.
\label{algo:prasayati_check}
\end{algorithm}

\subsection{Integration and Processing Pipeline}
The complete pipeline in Figure ~\ref{fig:chandassu_architecture} takes a raw padyam as input and returns a metrical assessment (Chandassu Score along with individual constraint scores). AksharamTokenizer segments the padyam into aksharam tokens, the LaghuvuGuruvu Generator labels each token as laghuvu (light) or guruvu (heavy), and PadyaBhedam Checker validates the resulting sequence against predefined rules for ganam, yati, prasa, and prasa yati, together with the aksharam-count and paadam-count constraints. The constraint scores are aggregated into the Chandassu Score. Error handling covers malformed input, incomplete padyams, and edge cases in aksharam token identification. Each module can be tested independently, and new prosodic constraints can be added without changing the other modules. The entire framework is implemented in the Python programming language \cite{10.5555/1593511}. 

\section{Evaluation Metrics}
\label{sec:evaluation_metrics}

We propose a comprehensive evaluation framework consisting of five fine-grained metrics that assess individual prosodic constraints and an aggregated Chandassu Score for overall metrical correctness:

\begin{enumerate}

\item \textit{n\_aksharaalu score} ($c_{na}$): The ratio of number of aksharam tokens present in given input text to the number of expected aksharam tokens in particular padyam.

\item \textit{n\_paadalu score} ($c_{np}$): The ratio of the number of paadams found in the given input text to the expected number of paadams in the padyam.
For instance, a line is considered to be paadam if at least one aksharam token is present during calculation of paadamwise gana kramam data.

\item \textit{gana\_kramam score} ($c_{gk}$): The ratio of total number of ganams matched sequentially to the total expected number of ganams for the padyam.

\item  \textit{yati score} ($c_{yt}$): The ratio of number of paadams of yati match to the total number of paadams expected to satisfy yati in the padyam. 

\item \textit{prasa score} ($c_{pr}$): The ratio of frequency of the modal prasa aksharam token to the expected number of paadams for the padyam.

\end{enumerate}

\begin{equation}
C = \frac{1}{n} \sum_{i=1}^{n} w_i*c_i,
\quad \text{where } c_i \text{ is each fine-grained constraint and } n \text{ is the number of constraints}
\label{eq:chandassu_score}
\end{equation}

Our proposed Chandassu Score ($C$) aggregates these individual constraints through arithmetic averaging as defined in Equation~\ref{eq:chandassu_score}, providing a unified quantitative measure of metrical correctness that ranges from 0 to 1, where higher values indicate greater adherence to traditional prosodic requirements for the corresponding padyam type. The Chandassu Score measures rule conformance, not classification accuracy, so it should be read as a benchmark on the curated corpus. This framework evaluates structural and prosodic correctness while assuming semantic validity of the input text; semantic evaluation could be integrated into future versions of the Chandassu Score. Also, this framework focuses on individual padyam assessment rather than collection-level metrics, as our primary objective is quantifying metrical correctness at the poem level. Future extensions could incorporate satakam-level evaluation through an additional \textit{n\_padyam score ($c_{nd}$)} measuring completeness of poetry collections, thereby expanding the framework's applicability to broader literary analysis tasks.

\section{Results and Discussion}

\begin{figure}[t!]
\begin{center}
\includegraphics[width=0.85\textwidth]{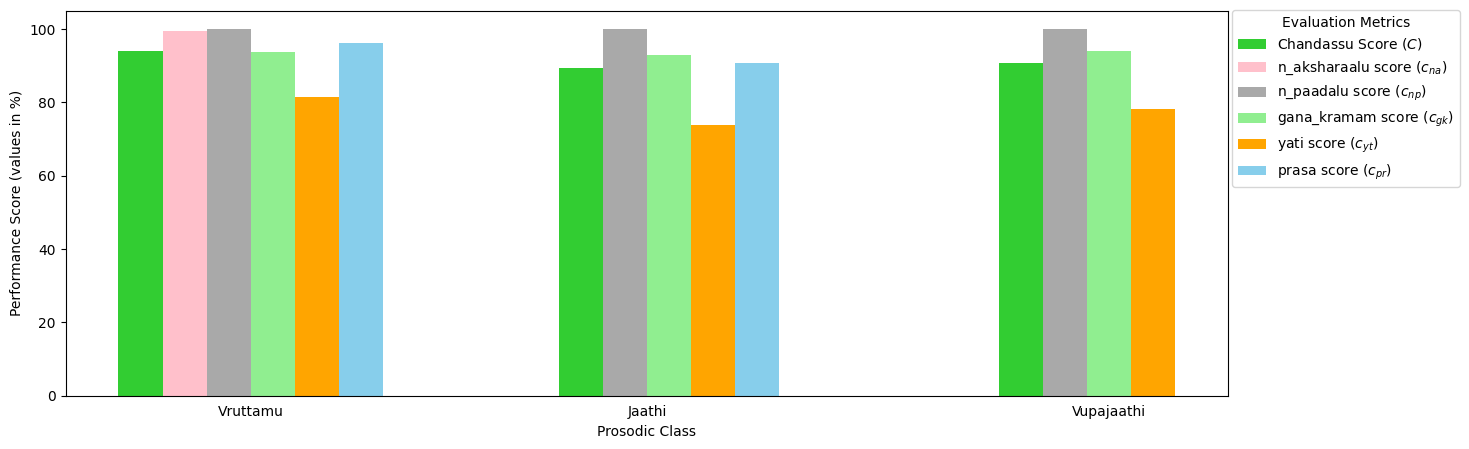}
\end{center}
\caption{Performance evaluation across prosodic classes}
\label{fig:evaluation_benchmark_class}
\end{figure}

\begin{figure}[t!]
\begin{center}
\includegraphics[width=0.85\textwidth]{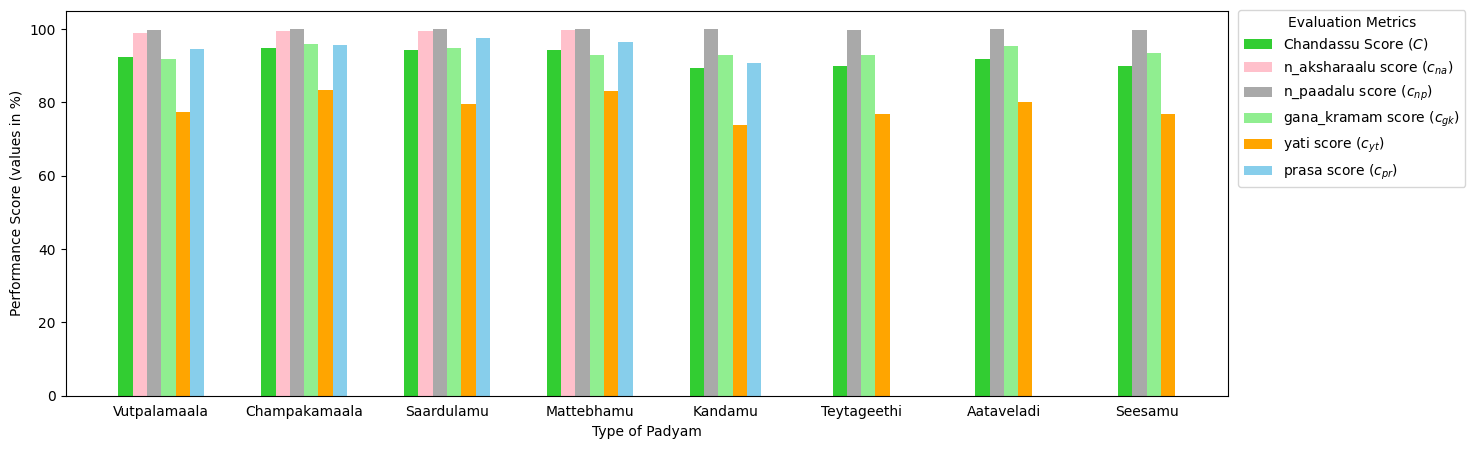}
\end{center}
\caption{Performance evaluation across padyam types}
\label{fig:evaluation_benchmark_type}
\end{figure}

\begin{table}[b!]
\centering
\begin{tabular}{lllllll}
\toprule
Prosodic Class& $c_{na}$& $c_{np}$& $c_{gk}$& $c_{yt}$& $c_{pr}$& $C$\\
\midrule
Vruttamu& \textbf{99.43}& 99.94& 93.78& \textbf{81.38}& \textbf{96.12}& \textbf{94.13} \\
Jaathi& -& \textbf{100.00}& 92.85& 73.94& 90.78& 89.39 \\
Vupajaathi& -& 99.91& \textbf{94.12}& 78.21& -& 90.75 \\
\bottomrule
\end{tabular}
\vspace{0.6em}
\caption{Performance evaluation by prosodic class (values in \%)}
\label{tab:evaluation_scores_classes_padyams}
\end{table}
 
\begin{table}[b!]
\centering
\begin{tabular}{lllllll}
\toprule
Padyam Type& $c_{na}$& $c_{np}$& $c_{gk}$& $c_{yt}$& $c_{pr}$& $C$ \\
\midrule
Vutpalamaala& 98.97& 99.70& 91.87& 77.36& 94.60& 92.50 \\
Champakamaala& 99.54& \textbf{100.00}& \textbf{96.02}& \textbf{83.48}& 95.69& \textbf{94.95} \\
Saardulamu& 99.35& \textbf{100.00}& 94.88& 79.66& \textbf{97.59}& 94.29 \\
Mattebhamu& \textbf{99.63}& \textbf{100.00}& 92.89& 83.02& 96.52& 94.41 \\
Kandamu& -& \textbf{100.00}& 92.85& 73.94& 90.78& 89.39 \\
Teytageethi& -& 99.85& 92.85& 76.81& -& 89.84 \\
Aataveladi& -& \textbf{100.00}& 95.43& 80.18& -& 91.87 \\
Seesamu& -& 99.83& 93.46& 76.71& -& 90.00 \\
\bottomrule
\end{tabular}
\vspace{0.6em}
\caption{Performance evaluation by padyam type (values in \%)}
\label{tab:evaluation_scores_types_padyams}
\end{table}

Our computational framework achieves strong performance across all proposed evaluation metrics, establishing the first quantitative benchmark for Telugu chandassu analysis. The algorithm attains an overall Chandassu Score of 91.73\% across all the poems present in the prepared dataset in Data Collection \ref{sec:DataCollection}, demonstrating robust metrical pattern recognition and validation capabilities. Individual metric performance reveals varying degrees of constraint satisfaction: structural metrics show exceptional accuracy with \textit{n\_aksharaalu score} ($c_{na}$) of 99.43\% and \textit{n\_paadalu score} ($c_{np}$) of 99.93\%, indicating near-perfect identification of syllabic count and line structure requirements. The \textit{gana\_kramam score} ($c_{gk}$) of 93.82\% reflects strong ganam sequence matching, while the \textit{prasa score} ($c_{pr}$) of 94.54\% demonstrates effective alliteration pattern recognition. The \textit{yati score}  ($c_{yt}$) of 78.69\% represents the most challenging constraint, reflecting the inherent complexity of caesura identification in traditional Telugu poetry. Laghuvu-Guruvu classification achieves 100\% accuracy against the aksharam token annotations across all 4,651 padyams. Performance analysis across prosodic classes and padyam types across each constraint, detailed in Table~\ref{tab:evaluation_scores_classes_padyams}, ~\ref{tab:evaluation_scores_types_padyams} and Figure ~\ref{fig:evaluation_benchmark_class}, ~\ref{fig:evaluation_benchmark_type} respectively, reveals systematic variations in metric performance. 

These variations reflect the differential constraint complexity inherent to each prosodic category: Jaathi and Vupajaathi classes lack aksharam count constraints (for the padyam types considered in this study), while Vupajaathi additionally omits prasa requirements (for the current types of padyams), resulting in modified evaluation profiles for these categories. The analyses by prosodic class and padyam type provide insights into the algorithm's effectiveness across the diverse landscape of Telugu metrical poetry and identifies specific areas for future algorithmic refinement. Errors in yati recognition arise mainly from sandhi. Sandhi merges adjacent words, so the word-initial letter at the yati position is no longer explicit. Handling sandhi explicitly, together with additional techniques, should therefore raise yati accuracy, and we identify this as the main direction for future work.

\section{Conclusion}
This work introduces the first comprehensive computational framework for Telugu chandassu analysis, comprising four key contributions: a curated dataset of 4,651 LaghuvuGuruvu-annotated padyams, AksharamTokenizer for prosodic-aware tokenization, an algorithmic approach for automated metrical analysis, and Chandassu Score for quantitative evaluation. Our algorithm achieves a benchmark Chandassu Score of 91.73\%, establishing foundational performance metrics for Telugu computational prosody. The framework addresses critical gaps in Telugu computational linguistics while providing potentially useful resources for literary analysis, digital humanities research, and prosodic pattern recognition. 

Beyond its technical contributions, the framework supports the communities that sustain Telugu chandassu. Poets and literary societies can use this framework for real-time metrical validation during composition, supporting the continued practice of chandassu as a living literary tradition. Educators can integrate it into Telugu language pedagogy, giving students immediate, rule-based feedback on ganam, yati, and prasa. Literary scholars and heritage initiatives gain quantitative tools for the systematic analysis, verification, and digital archiving of historical poetry corpora, and the openly licensed dataset provides a durable, community-accessible record of this knowledge. Computational linguists benefit from standardized resources for prosodic research. Future work will extend the system to handle sandhi phenomena, enabling more sophisticated phonological analysis and broader applicability to classical Telugu texts, and will engage poets, educators, and cultural institutions directly in evaluating and refining the framework.

\section{Acknowledgements}
We acknowledge Sesha Sai Vadapalli and Kalepu Nagabhushana Rao for their invaluable contributions in curating and maintaining the extensive Telugu literature collection on andhrabharati.com, which served as the primary data source for this research. We thank Appana Mohan Naga Phani Kumar for proofreading this article. We express sincere gratitude to our parents and family members for their continuous support throughout this work. This work was undertaken with the grace of Sri Ramalinga Chowdeswari Devi.

\bibliographystyle{plain}

\end{document}